\documentclass[preprint,11pt]{elsarticle}
\UseRawInputEncoding
\usepackage{amsmath, amssymb}
\usepackage{graphicx}
\usepackage{hyperref}
\usepackage{booktabs}
\usepackage{xcolor}
\usepackage{tablefootnote}
\usepackage[utf8]{inputenc}
\journal{Artificial Intelligence}
\usepackage{xcolor}
\usepackage{amsfonts}
\usepackage{graphicx} 
\usepackage[a4paper, total={6in, 8in}]{geometry}

\date{June 2025}
\usepackage{float}
\usepackage{svg}
\usepackage{pifont}
\usepackage{booktabs}
\usepackage{tabularx}
\usepackage{pifont}
\usepackage{amsmath}
\usepackage[linesnumbered,ruled,vlined]{algorithm2e}
\SetKw{KwParams}{Parameters}


\begin{document}

\begin{frontmatter}

\title{Transformer World Model for Sample Efficient Multi-Agent Reinforcement Learning}
\author[inst1]{Azad Deihim}
\author[inst1]{Eduardo Alonso}
\author[inst2]{Dimitra Apostolopoulou}
\address[inst1]{Artificial Intelligence Research Centre (CitAI), City St George’s, University of London}
\address[inst2]{Oxford Institute for Energy Studies (OIES), University of Oxford}

\begin{abstract}
We present the Multi-Agent Transformer World Model (MATWM), a novel transformer-based world model designed for multi-agent reinforcement learning in both vector- and image-based environments. MATWM combines a decentralized imagination framework with a semi-centralized critic and a teammate prediction module, enabling agents to model and anticipate the behavior of others under partial observability. To address non-stationarity, we incorporate a prioritized replay mechanism that trains the world model on recent experiences, allowing it to adapt to agents' evolving policies. We evaluated MATWM on a broad suite of benchmarks, including the StarCraft Multi-Agent Challenge, PettingZoo, and MeltingPot. MATWM achieves state-of-the-art performance, outperforming both model-free and prior world model approaches, while demonstrating strong sample efficiency, achieving near-optimal performance in as few as 50K environment interactions. Ablation studies confirm the impact of each component, with substantial gains in coordination-heavy tasks.
\end{abstract}


\end{frontmatter}

\section{Introduction}
Reinforcement learning (RL) is a framework in which agents learn to make decisions by interacting with an environment to maximize cumulative rewards, typically through trial and error. Deep RL (DRL) extends this by using deep neural networks to approximate policies and value functions, enabling agents to operate in high-dimensional or complex environments. Within DRL, model-free algorithms refer to methods that learn directly from experience without constructing an explicit model of the environment’s dynamics. These include popular approaches like Q-learning and policy gradients. Model-free DRL methods have led to breakthroughs in many DRL settings \cite{coma, maddpg, mappo, Qmix, QPLEX, qtran, VDN}, but they often suffer from sample inefficiency, requiring millions of interactions to learn effective behaviors, making them impractical in environments where data collection is expensive or risky \cite{Hernandez-Leal}. This efficiency gap has limited the use of RL in real-world tasks where large-scale interaction is not feasible. World models, however, offer a promising solution to this issue. In RL, a world model is a learned representation of the environment that allows agents to simulate and learn from predicted future outcomes without interacting with the real environment \cite{worldmodelreview}. By enabling agents to "imagine" future trajectories given some contextual observations and actions, they can compute and learn from up to 16 imagined future steps per real sample. A world model typically uses neural networks to replicate environment dynamics like transitions and rewards. This capability dramatically improves sample efficiency, which can make DRL more attractive for domains where real-world sampling of millions of experiences is impractical, such as robotics, healthcare, or autonomous driving systems. 

While world models have demonstrated substantial improvements in sample efficiency for single-agent RL, their application to multi-agent environments remains largely unexplored due to the added complexity of modeling interactions among agents. Only a few existing frameworks, such as MBVD~\cite{MBVD}, MAMBA~\cite{MAMBA}, and MARIE~\cite{marie}, have attempted to bring world modeling into multi-agent RL (MARL), and although these models achieve promising results, they still fall short of matching the complexity needed to tackle environments on par with those addressed by their single-agent counterparts, such as vast 3D environments like Minecraft \cite{minerl}.

In multi-agent settings, accurately predicting long-horizon trajectories becomes increasingly difficult, as the interactions of multiple agents introduce additional sources of error that can compound and propagate throughout the entire imagination rollout. As a result, horizon lengths that are feasible in single-agent settings often become infeasible in multi-agent environments. A further challenge stems from two strategies for modeling multi-agent dynamics: centralized and decentralized. A centralized world model approach may better capture agent interdependencies by using all agent information to compute a joint trajectory, but it scales poorly as the number of agents grows, leading to increased spatial complexity. A decentralized world model approach that only computes local trajectories based on local information, while more scalable, suffers from non-stationarity issues due to unpredictable and constantly evolving interactions with other agents, resulting in an inaccurate representation of imagined state transitions. However, a major advantage of using world models in multi-agent systems is the ability to apply new dynamics to old experiences. Generally, learning from older experiences has the potential to become harmful because they reflect outdated behaviors from other agents, but a world model with the capacity to track and update with each agent’s evolving behavior can reinterpret old experiences given new dynamics, making old transitions more relevant.

To advance the capabilities of multi-agent world models, we introduce a novel Multi-Agent Transformer World Model (MATWM)\footnote{github.com/azaddeihim/matwm} framework, built upon and inspired by STORM \cite{storm}, a leading transformer world model architecture for single-agent systems. We selected STORM as the foundation for MATWM because it consolidates a wide range of best practices and architectural advances found across leading single-agent world models \cite{dreamerv2, dreamerv3, twm}, such as transformer-based sequence modeling \cite{transformer}, Kullback–Leibler (KL) balance and free bits \cite{freebits}, percentile return normalization, two-hot symlog reward targets, discrete latent representations, and many more --- several of which have not yet been incorporated into any existing multi-agent world model architecture. This makes it an ideal starting point for developing a more complex and capable multi-agent world model, allowing us to focus our contributions on the novel extensions required for multi-agent imagination and coordination, such as teammate prediction, prioritized replay sampling, and action space scaling to distinguish between agents. We utilize a decentralized world model approach, and our agent training is semi-centralized, in that it does not explicitly use information from other agents during training, but instead uses imagined behavior of other agents during training. This follows the Centralized Training and Decentralized Execution (CTDE) design, but bypasses the need to have access to all agents' information during agent training. 

The contributions of MATWM are as follows:
\begin{enumerate}
    \item Solve complex multi-agent benchmarks, such as the StarCraft Multi-Agent Challenge (SMAC), PettingZoo, and MeltingPot, with as few as 50K real environment interactions, which, to our knowledge, is the lowest sample allowance of any existing MARL algorithm.
    \item We incorporate several methods from top-performing single-agent world model architectures that have not yet been adopted in multi-agent settings, including but not limited to free bits,  percentile return normalization, and two-hot symlog reward targets \cite{dreamerv3, storm}.

    \item We design a lightweight and effective teammate predictor module for our world model to model agents' behavior to improve cooperation between agents, where communication is otherwise not possible. 
    \item We employ a prioritized replay buffer to ensure the world model is trained on more recent and informative experiences, enabling it to stay aligned with the agents' evolving policies. Additionally, we use an action scaling mechanism to encode agent-specific information, helping the world model differentiate between individual agents.

    \item We present, to our knowledge, the first multi-agent world model capable of learning from image-based observations. Unlike prior work, which has been limited to low-dimensional vector inputs, MATWM supports visual environments and demonstrates strong performance in this setting.
\end{enumerate}

The remainder of this paper is structured as follows. In Section~\ref{sec:related_work}, we review prior work on world models and their extension to MARL. Section~\ref{sec:method} details the architecture and training process of MATWM, including our key design choices and novel components. Section~\ref{sec:experiments} presents experimental results across SMAC, PettingZoo, and MeltingPot environments, demonstrating the effectiveness and efficiency of our approach and analyzing the contribution of individual components via ablation studies. Finally, Section~\ref{sec:conclusion} summarizes our contributions and outlines directions for future work.

\section{Related Work}
\label{sec:related_work}
World models have shown significant promise in improving sample efficiency through imagination-based training \cite{worldmodelreview, worldmodelreviewFeng, worldmodelreviewFMLou}. Recent efforts proposed to improve sample efficiency in model-free algorithms have been successful through the use of auxiliary objectives \cite{kim2023sampleMF, liu2025learningMF} or data augmentation \cite{ma2023dataaug, liu2023dataaug}. However, even with these improvements, model-free algorithms still require significantly more training samples to achieve results similar to those of world model-based algorithms. 

The original world model \cite{worldmodel} pioneered the idea in the single-agent setting, learning models for imagining future trajectories using autoencoders and recurrent neural networks (RNN)  \cite{Rnn}. Following this, several improvements were made in imagined state representations, agent training, and structure and contents of the world model through frameworks such as SimPLE \cite{simple} and the Dreamer series \cite{dreamerv1, dreamerv2, dreamerv3}. Following the debut of Dreamer, several new world model architectures, IRIS \cite{iris}, TWM \cite{twm}, TransDreamer \cite{transdreamer}, and STORM \cite{storm}, were able to further enhance the Dreamer method with the use of transformers as the core sequence model rather than GRU\cite{GRU} or other RNN-based architectures, as transformers\cite{transformer} typically outperform RNNs in a wide variety of sequence modeling tasks due to their ability to model long-range dependencies and enable parallel computation. However, all of these world model architectures only considered single-agent systems and would likely require significant rework to enable success in multi-agent settings. 

The existing literature on multi-agent world models is very limited. MAZero \cite{mazero}, a multi-agent extension of MuZero \cite{muzero}, is a model-based approach to MARL that closely resembles world models; it uses Monte Carlo Tree Search to plan future trajectories instead of a traditional neural network-based world model, which was computationally expensive and scaled poorly with an increased number of agents or larger action spaces. MBVD and MAMBA were among the first model-based approaches to employ neural network-based world models in MARL. MBVD leverages shared latent rollouts and value decomposition to facilitate coordination, utilizing a GRU-based sequence model \cite{GRU} to generate imagined trajectories. While it introduced an important architectural step forward as one of the earliest world models tailored for multi-agent systems, it still suffers from low sample efficiency, often requiring over one million environment steps to reach performance levels that more recent multi-agent world model methods can achieve in under 100,000 steps. MAMBA, a multi-agent extension of DreamerV2, introduces a centralized world model approach that uses joint-agent dynamics to compute a joint-agent state using a GRU-based sequence model, enabling imagination-based learning in partially observable environments. It has demonstrated strong performance across a range of challenging MARL tasks, including those with high agent counts and partial observability, outperforming traditional model-free approaches and earlier model-based methods. While effective for modelling interactions between agents, a downside of the centralized world model can become extremely computationally expensive as the number of agents grows.

At the time of writing, MARIE \cite{marie} introduces the first and only Transformer-based world model for multi-agent systems, which achieved a substantial improvement over MBVD and a marginal improvement MAMBA in a wide range of SMAC maps. It learns decentralized local dynamics alongside centralized feature aggregation via a Perceiver \cite{Perceiver}. The aggregation module is used to compute contextual information about the entire team from a given agent’s perspective. These aggregated features are inserted into each agent’s local token sequence and passed to the shared dynamics model, allowing for coordination and mitigating non-stationarity during imagination. This method proved successful in enabling multi-agent imagination while being far more memory-efficient than the centralized world model approach found in MAMBA. Our approach shares the goal of enabling sample efficient multi-agent learning through imagination but takes a complementary route: instead of modeling all agent interactions through centralized aggregation, we explicitly model other agents' behavior with a learned teammate predictor. The computational cost of the teammate predictor remains nearly constant as the number of agents increases, whereas the aggregation module's computational cost does increase with the number of agents.

MARIE, along with MAMBA, use PPO-style training \cite{ppo} for their world model and agents; this can be advantageous in multi-agent systems as it ensures that experiences being used to train the world model and agents are recent and do not reflect outdated agent behaviours, but PPO-style training is generally not as sample efficient due to infrequent updates and can suffer from catastrophic forgetting \cite{kaplanis2019policy}. Instead, we adopt a DreamerV3-style \cite{dreamerv3} training framework augmented with a prioritized replay mechanism that emphasizes recent experiences. This setup enables more frequent updates, training both the world model and the agent after each environment step. We hypothesize that this approach enhances sample efficiency due to the higher volume of updates, while mitigating overfitting in agent training by continuously updating imagined trajectories in sync with evolving policies. Additionally, we extend our framework to support both image-based and vector-based observations, whereas prior multi-agent world models have been limited to vector-based inputs only.

\subsection{Latent Representations}

One of the primary distinctions between world model implementations lies in how state information is represented and provided to the agent. Training directly on raw observations can introduce significant noise and dramatically increase computational requirements \cite{efficientzero, dreamerv2, twm}, especially so for image data. To address this, most world models now rely on learned latent representations to encode essential features while discarding irrelevant noise and decreasing the dimensionality of the data. Early world model approaches typically employed continuous latent spaces, most commonly learned via standard variational autoencoders (VAEs) \cite{vae}. As opposed to the Vector Quantized Variational Autoencoder (VQ-VAE), which is a self-supervised method that learns discrete latent representations, achieving comparable performance while producing significantly more compact encodings \cite{vqvae}; for example, when applied to environments like DeepMind Lab \cite{deepmindlab}, they can represent observations using as few as 27 bits. Furthermore, empirical studies have shown that discrete latent spaces often outperform continuous ones on a wide range of RL tasks, even in model-free algorithms \cite{meyer2024harnessing}.

Nonetheless, DreamerV1~\cite{dreamerv1} and World Models~\cite{worldmodel} adopted continuous latent spaces for their frameworks. SimPLE \cite{simple} was among the first to instead utilize a VQ-VAE. However, this approach was soon surpassed by DreamerV2~\cite{dreamerv2}, which adopted a Categorical-VAE \cite{catvae} that achieved substantially better performance. Today, several of the most competitive single-agent world models employ Categorical-VAEs as their default. In contrast, Categorical-VAEs are less prevalent in multi-agent world models. MBVD utilizes continuous latent spaces, whereas MARIE employs discrete representations through a VQ-VAE. MAMBA, however, does use a Categorical-VAE. We hypothesize that our use of a Categorical-VAE will provide a slight representational advantage over the existing multi-agent world models that have not yet adopted it.

We present a comparative overview of MARIE, MAMBA, MBVD, and MATWM in Table~\ref{tab:wm_comparison}, highlighting key distinctions in architectural and training design. 
\begin{table}[h]
\centering
\scriptsize
\caption{Comparison of MATWM with leading multi-agent world model architectures}
\label{tab:wm_comparison}
\begin{tabularx}{\textwidth}{lXXXX}
\toprule
\textbf{Aspect} & \textbf{MARIE} & \textbf{MAMBA} & \textbf{MBVD} & \textbf{MATWM} \\
\midrule
\textbf{Multi-Agent Design}             & Yes & Yes & Yes & Yes \\
\textbf{Backbone Architecture}          & Transformer & GRU  & GRU & Transformer \\
\textbf{Latent Representation}         & VQ-VAE & Categorical VAE & VAE & Categorical VAE \\
\textbf{Critic Type}                   & Centralized & Centralized & Centralized  & Semi-centralized \\
\textbf{World Model Type}              & Hybrid & Centralized & Centralized & Decentralized \\
\textbf{Visual Observation Support}    & No & No & No & Yes \\
\textbf{Agent Training Strategy}       & PPO-style & PPO-style & Deep Q-learning & DreamerV3-style \\
\bottomrule
\end{tabularx}
\end{table}

\section{Method}
\label{sec:method}

All agents in the environment utilize a shared world model, which is trained using an equal distribution of experiences from each agent. In this paper, the focal agent refers to the agent currently being trained, evaluated, or associated with a given experience, or for whom the world model is generating predictions and trajectories. Our approach builds upon STORM~\cite{storm}, a single-agent world model architecture that integrates several state-of-the-art design principles, including transformer-based sequence modeling, symlog reward scaling, KL balance and free bits, and categorical latent representations. MATWM retains STORM's core components but adds key extensions for multi-agent settings. Specifically, MATWM adds a teammate predictor to model other agents’ behavior, a prioritized replay sampling strategy to maintain up-to-date world model training, and action scaling mechanisms that allow agents to be distinguished within shared environments. These additions enable MATWM to support multi-agent coordination without requiring centralized training or communication.
The framework utilized in this paper is consistent with that of other world model-based algorithms; agents learn a policy via imaginations generated by the world model. This is done via the following three steps that repeat in order:
\begin{enumerate}
    \item Populate a replay buffer with real experiences using the current policies of the agents.
    \item Train the world model using samples from the replay buffer.
    \item Sample real experiences from the replay buffer, compute imagined trajectories using the real experiences as a starting point, and update the agents' policies using the imagined trajectories. 
\end{enumerate}

These steps are iterated until the maximum number of allowed steps through the real environment is achieved. Each entry in the replay buffer contains an observation $o_t$, an action $a_t$, a reward $r_t$, a continuation flag $c_t$ (a Boolean variable denoting whether an episode has terminated), and an action mask $m_t$, if applicable. All agents will have their own replay buffer. Agents' actions will be randomly sampled until the replay buffer reaches 1,000 samples, after which agents and the world model will begin training, and agents will use their learned policy to decide actions. After each real environment step, the world model and the agents will undergo one epoch of training. For world model training, we sample a batch of 64-length sequences from the replay buffer, prioritizing recent samples using an exponentially decaying weighting system. We also ensure that there is no overlap between sequences to promote diversity in the training batch. When sampling from the replay buffer for agent training, we select experiences entirely at random without any weighting. We utilize a novel trick for agent-world model interactions by scaling each agent's action space to be mutually orthogonal. For instance, if one agent's discrete action space is \{0, 1, 2\}, the next agent's actions are offset to \{3, 4, 5\}, and so on for additional agents. This enables the world model to distinguish agents without requiring explicit IDs or embeddings.

\subsection{World Model}

\begin{figure}[t]
  \centering
  \includegraphics[trim=0cm 0cm 0cm 0cm, width=\linewidth]{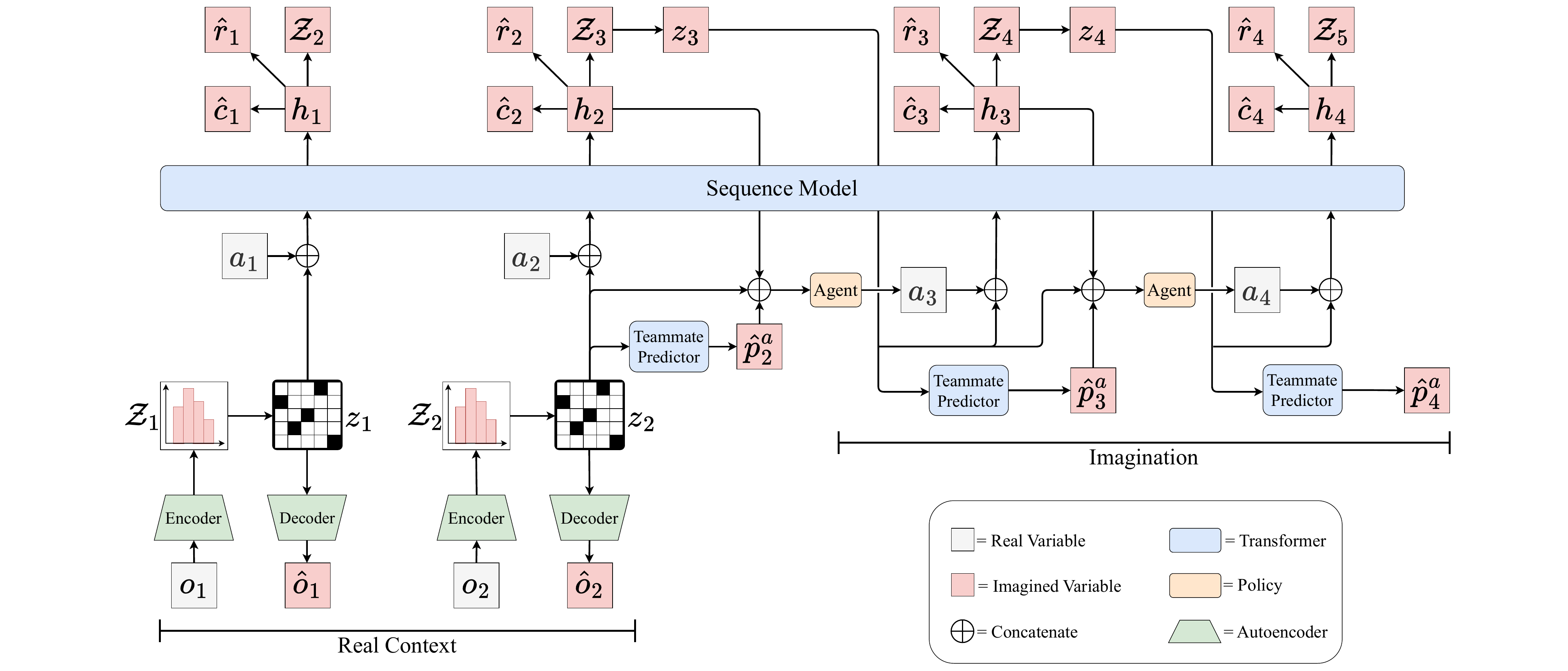}
  \caption{Overview of the MATWM architecture. The figure illustrates how real context (past observations and actions) are encoded and used as a starting point for imagination rollouts. Each agent encodes its local observation into a discrete latent state via an autoencoder. These latent states, combined with actions, are processed through an action mixer and a transformer-based sequence model to produce hidden states. The world model then predicts the next latent state, reward, continuation, and optionally, the action mask. The agent then uses the new latent state, predicted teammate action logits and the hidden state to compute a new action. The new action and latent state are then used to repeat this process.}
  \label{matwm}
\end{figure}

The world model is a collection of predictive models that can replicate the dynamics of the real environment, allowing the agents to interact with and learn from the world model, rather than the real environment directly. Our world model architecture comprises several modules that enable the accurate replication of the real environment's dynamics. Figure \ref{matwm} shows a diagram of the MATWM architecture and how a focal agent interacts with it, further explained in Algorithm \ref{alg:multiagent_imagination} found in \ref{imaginationalgo}. We build a world model that is shared by all agents, but only computes the focal agents trajectory using only the focal agent's information. 

We use an autoencoder to transform each observation \( o_t \) into a stochastic latent categorical distribution \( \mathcal{Z}_t \). Following prior work~\cite{dreamerv2, storm, dreamerv3}, we define \( \mathcal{Z}_t \) as consisting of 32 categorical variables, each with 32 discrete classes. The encoder \( q_\phi \) and decoder \( p_\phi \) are implemented as convolutional neural networks (CNNs) \cite{cnn} for image observations and multi-layer perceptrons (MLPs) for vector observations. The latent state \( z_t \) is sampled from \( \mathcal{Z}_t \) to represent the original observation \( o_t \). Since sampling from a categorical distribution is non-differentiable, we use the straight-through gradient estimator \cite{gradestimate} to preserve gradients for backpropagation. We define the encoder and decoder as follows:

\begin{equation}
\begin{aligned}
& \text { Encoder: } &z_t \sim q_\phi\left(z_t \mid o_t\right)=\mathcal{Z}_t \\
& \text { Decoder: } &\hat{o}_t=p_\phi\left(z_t\right) \\
\end{aligned}
\end{equation}

We utilize an action mixer, a single linear transformation denoted by the function $m_\phi(\cdot)$, that merges the latent state $z_t$ with the agent's action $a_t$, returning an embedded representation $e_t$. This is passed into a vanilla transformer \cite{transformer}, whose function is denoted by $f_\phi(\cdot)$, acting as the sequence model, which outputs the hidden states $h_{0: T}$ for all $e_{0: T}$. Then, the dynamics predictor, an MLP-based module, denoted by the function $g_\phi(\cdot)$, predicts the next latent state $\hat{z}_{t+1}$ from the $h_t$. Since no real observations are available during imagination, the dynamics predictor is critical for rolling out future states, as we would not be able to use the encoder to compute them. The reward and continuation predictors estimate rewards and continuation flags during imagination. We define these world model components as:

\begin{equation}
\begin{aligned}
& \text { Action mixer: }  &e_t =m_\phi^E\left(z_t, {a}_{t}\right) \\
& \text { Sequence model: }  &h_{0: T} =f_\phi\left(e_{0: T}\right) \\
& \text { Dynamics predictor: }  &\hat{\mathcal{Z}}_{t+1} =g_\phi^D\left(\hat{z}_{t+1} \mid h_t\right) \\
& \text { Teammate predictor: } & \hat{p}_{t,0}^{(a)}, \cdot \cdot \cdot, \hat{p}_{t,N}^{(a)}=f_\phi(z_{0:T}) \\
& \text { Reward predictor: } &\quad \hat{r}_t=g_\phi^R\left(h_t\right) \\
& \text { Continuation predictor: } &\quad \hat{c}_t=g_\phi^C\left(h_t\right) \\
& \text { Action Mask predictor: } &\quad \hat{m}_t=g_\phi^M\left(z_t\right)
\end{aligned}
\end{equation}

To account for the behaviors of other agents and enable cooperative behavior during both imagination and interaction with the real environment, the teammate predictor utilizes a transformer to infer the actions of all $N$ non-focal agents based on the focal agent's latent state history.  Lastly, the action mask predictor is critical in partially observable or constrained environments like SMAC, where agents are only allowed to execute legal actions. Since imagined rollouts do not have access to real action masks, we train a predictor to estimate them. This allows us to avoid degenerate scenarios where an agent conducts an illegal action, for which the sequence model may produce an invalid state, thus tainting the entire imagination rollout.

For each agent in the environment, we randomly sample transitions from their replay buffer, then use them to update the world model via the total world model loss function:
\begin{equation}
\mathcal{L}(\phi)=\frac{1}{B T} \sum_{n=1}^B \sum_{t=1}^T\left[\mathcal{L}_t^{\mathrm{rec}}(\phi)+\mathcal{L}_t^{\mathrm{rew}}(\phi)+\mathcal{L}_t^{\mathrm{con}}(\phi)+ \mathcal{L}_t^{\mathrm{team}}(\phi)+ \beta_1(\mathcal{L}_t^{\mathrm{mask}}(\phi) +   \mathcal{L}_t^{\mathrm{dyn}}(\phi))+\beta_2 \mathcal{L}_t^{\mathrm{rep}}(\phi)\right],
\end{equation}
where $B$ is the batch size, $T$ is the batch length, and $\mathcal{L}_t^{\mathrm{rec}}$, $\mathcal{L}_t^{\mathrm{rew}}$, $\mathcal{L}_t^{\mathrm{con}}$, $\mathcal{L}_t^{\mathrm{team}}$, $\mathcal{L}_t^{\mathrm{mask}}$, $\mathcal{L}_t^{\mathrm{dyn}}$, and $\mathcal{L}_t^{\mathrm{rep}}$ are sub-loss functions for different subsets of the world models components.

The reconstruction loss, $\mathcal{L}_t^{\text {rec }}$, is represented by the mean-squared error between the real observation and the predicted reconstruction of the observation, facilitating the training of the encoder and the decoder:
\begin{equation}
\mathcal{L}_t^{\text {rec }}(\phi) =(\hat{o}_t-o_t)^2.
\end{equation}

\noindent Next, $\mathcal{L}_t^{\text {rew }}$ and $\mathcal{L}_t^{\text {con }}$ are the components of the world model loss that facilitate the training of the reward and continue predictors, respectively. In $\mathcal{L}_t^{\text {rew }}$, we utilize a symlog two-hot loss, as described in \cite{dreamerv3}. $\mathcal{L}_t^{\text {con }}$ is the binary cross-entropy loss between the ground truth continuation flag $c_t$ and predicted continuation flag $\hat{c}_t$. These sub-loss functions are computed as follows:

\begin{align}
&\mathcal{L}_t^{\text {rew }}(\phi) = \mathcal{L}^{\text {sym }}\left(\hat{r}_t, r_t\right), \\
&\mathcal{L}_t^{\text {con }}(\phi) = c_t \log \hat{c}_t+\left(1-c_t\right) \log \left(1-\hat{c}_t\right).
\end{align}

\noindent The action mask loss $\mathcal{L}_t^{\text {mask }}$ is computed via the binary cross-entropy between the predicted action mask $\hat{m}_t$ and the ground-truth binary action mask $m_t$:

\begin{equation}
\mathcal{L}_t^{\text {mask }}(\phi) = m_t \log \hat{m}_t+\left(1-m_t\right) \log \left(1-\hat{m}_t\right).
\end{equation}

\noindent The teammate predictor, given the latent state sequence of the focal agent, outputs a logits representing the probability of each teammate’s possible actions. The ground truth teammate actions are used to compute a standard cross-entropy loss. For each teammate $i \in {0, \cdot \cdot \cdot, N}$ and each possible action $a \in {0, \cdot \cdot \cdot, A}$, where $A$ is the number of discrete actions, we define:

\begin{equation}
\mathcal{L}_t^{\text{team}}(\phi) = -\sum_{i=1}^{N} \sum_{a=1}^{A} \delta(a_{t,i} = a) \log \hat{p}_{t,i}^{(a)},
\end{equation}

\noindent where $a_{t,i}$ is the ground truth action of teammate $i$ at time $t$, and $\hat{p}_{t,i}^{(a)}$ is the predicted probability of teammate $i$ selecting action $a$. During training, the latent state input to the teammate predictor is given as $\operatorname{sg}(z_{0:T})$, where $\operatorname{sg}(\cdot)$ denotes the stop-gradient operator. This is important as the actions can be noisy and random due to undertraining and/or entropy, and we do not want this random noise to backpropagate to the encoder.

The sub-loss functions $\mathcal{L}_t^{\mathrm{dyn}}$ and $\mathcal{L}_t^{\mathrm{rep}}$ are both represented as Kullback–Leibler (KL) divergences, and only differ in the placement of the stop-gradient operator, $\operatorname{sg}(\cdot)$. Here, $\mathcal{L}_t^{\mathrm{dyn}}$ aims to train the sequence model to predict the next latent state accurately. However, $\mathcal{L}_t^{\mathrm{rep}}$ incentivizes the encoder to compute latent states that are easier to predict correctly. We follow \cite{dreamerv3, storm} in employing free bits \cite{freebits} for these sub-loss functions to effectively disable them when they are sufficiently minimized, allowing the main objective loss function to focus on other sub-losses. This also mitigates trivial dynamics that are easy to predict but lack sufficient input information. These sub-losses are given by:

\begin{subequations} \label{eq:loss_terms}
\begin{align}
\mathcal{L}_t^{\mathrm{dyn}}(\phi) &= \max \left(1, \mathrm{KL}\left[\operatorname{sg}\left(q_\phi\left(z_{t+1} \mid o_{t+1}\right)\right) \| g_\phi^D\left(\hat{z}_{t+1} \mid h_t\right)\right]\right), \label{eq:loss_dyn} \\
\mathcal{L}_t^{\mathrm{rep}}(\phi) &= \max \left(1, \mathrm{KL}\left[q_\phi\left(z_{t+1} \mid o_{t+1}\right) \| \operatorname{sg}\left(g_\phi^D\left(\hat{z}_{t+1} \mid h_t\right)\right)\right]\right). \label{eq:loss_rep}
\end{align}
\end{subequations}

\subsection{Training Structure}
To train agents using imagined experience, we start from a context window of real observations and actions. The world model uses this context to produce an initial latent state and hidden state. During each imagination step, the focal agent samples an action based on its current state and the predicted action distribution of its teammates, produced by the teammate predictor. When enabled, the action mask predictor is used to enforce action feasibility constraints. The action is then passed to the world model to predict the next latent state, hidden state, reward, and termination signal, which are stored in temporary rollout buffers. The resulting trajectories, stored in that temporary rollout buffer, allow the agent policies to learn entirely from imagination, without requiring new environment interactions. 

Each agent is trained in isolation via a shared decentralized world model. Agents do not interact with one another directly; instead, they interact with imagined versions of other agents whose behavior is captured within the state information and by the teammate predictor. The rationale behind this approach is that the world model will treat non-focal agents as it would any other non-deterministic entity, similar to the non-agent adversaries in a single-agent Atari game, predicting their behavior and thus impact on the following state(s) to the best of its ability. The alternative, having agents coexist in and interact with the same imagined environment, poses several challenges. Firstly, the hidden state for each agent $h_t$ may encompass information from all agents, potentially leading to situations where agents receive and become reliant on information they are not meant to have, especially in partially observable environments. Secondly, computing a joint trajectory would cause the memory requirements of the world model to scale poorly with the number of agents, making it impractical for environments involving many agents. Nonetheless, we hypothesize that our training structure could exacerbate the non-stationarity issue, as the world model might be displaying outdated behaviors of the agents during imagination. Again, we alleviate this issue by using a replay memory structure that prioritizes sampling recent memories when training the world model. We share the full training algorithm in \ref{matwmalgo}.

The state $s_t$, for each agent, is formed by concatenating the predicted action logits $\hat{p}_{t}^{(a)}$ of all non-focal agents and its own $z_t$ and $h_t$. Agents are trained through an actor-critic learning algorithm. Each agent will have its own critic, which only takes in local state $s_t$ of the critic's respective agent along with that agent's $a_t$. A centralized critic would not be feasible here because during imagination, the agents' states are very likely to become desynchronized, so the actions conducted by an agent in its imagined state may not be transferable to another agent's imagined state, even in the same time step; additionally, they may have imaged states that are conflicting and cannot legally exist at the same time. This is why we use the teammate predictor to predict what the non-focal agents would do in the focal agent's current state, allowing us to have a semi-centralized critic that does not require having direct access to non-focal agent information. The Critic and Actor are denoted by: 

\begin{equation}
\begin{aligned}
&\text{Critic}: \quad V_\psi\left(s_t\right) \approx \mathbb{E}_{\pi_\theta, p_\phi}\left[\sum_{k=0}^{\infty} \gamma^k r_{t+k}\right], \\
&\text{Actor}: \quad a_t \sim \pi_\theta\left(a_t \mid s_t\right).
\end{aligned}
\end{equation}

\begin{equation}
\begin{aligned}
\mathcal{L}(\theta) & =\frac{1}{B L} \sum_{n=1}^B \sum_{t=1}^L\left[-\operatorname{sg}\left(\frac{G_t^\lambda-V_\psi\left(s_t\right)}{\max (1, S)}\right) \ln \pi_\theta\left(a_t \mid s_t\right)-\eta H\left(\pi_\theta\left(a_t \mid s_t\right)\right)\right], \\
\mathcal{L}(\psi) & =\frac{1}{B L} \sum_{n=1}^B \sum_{t=1}^L\left[\left(V_\psi\left(s_t\right)-\operatorname{sg}\left(G_t^\lambda\right)\right)^2+\left(V_\psi\left(s_t\right)-\operatorname{sg}\left(V_{\psi^{\mathrm{EMA}}}\left(s_t\right)\right)\right)^2\right].
\end{aligned}
\end{equation}

\noindent Where $H(\cdot)$ represents the entropy of the policy distribution, and the constants $\eta$ and $L$ denote the entropy regularization coefficient and the imagination horizon, respectively. The $\lambda$-return $G_t^\lambda$ is recursively defined \cite{rl} over the imagination horizon  as follows:

\begin{equation}
\begin{aligned}
G_t^\lambda & \doteq r_t+\gamma c_t\left[(1-\lambda) V_\psi\left(s_{t+1}\right)+\lambda G_{t+1}^\lambda\right], \\
G_L^\lambda & \doteq V_\psi\left(s_L\right).
\end{aligned}
\end{equation}

The normalization factor $S$ used in the actor loss (Equation 11) is defined in Equation (13). It is computed as the difference between the 95th and 5th percentiles of the $\lambda$-return $G_t^\lambda$ within the batch \cite{dreamerv3}:

\begin{equation}
S=\operatorname{percentile}\left(G_t^\lambda, 95\right)-\operatorname{percentile}\left(G_t^\lambda, 5\right).
\end{equation}

To regularize the value function, we maintain an exponential moving average (EMA) of the critic parameters $\psi$. As defined in Equation (14), $\psi_t$ represents the current critic parameters, $\sigma$ is the decay rate, and $\psi_{t+1}^{\text{EMA}}$ denotes the updated EMA. This technique helps stabilize training and mitigate overfitting:

\begin{equation}
\psi_{t+1}^{\mathrm{EMA}}=\sigma \psi_t^{\mathrm{EMA}}+(1-\sigma) \psi_t.
\end{equation}
\section{Experiments}
\label{sec:experiments}

We evaluate MATWM on three benchmark suites: the StarCraft Multi-Agent Challenge\cite{smac} for vector-based environments, and PettingZoo \cite{pettingzoo} and MeltingPot\cite{meltingpot} for image-based, cooperative multi-agent settings. SMAC provides partially observable micro-management scenarios in StarCraft II, where agents must coordinate to defeat built-in AI opponents. It includes scenarios (maps) of varying difficulty, with \textit{easy} maps requiring basic coordination and \textit{hard} maps generally involving tighter coordination and complex tactics. PettingZoo offers diverse multi-agent environments; we focus on PettingZoo Butterfly as those are both image-based and cooperative. MeltingPot targets social interaction, with visual environments that require cooperation, competition, and negotiation among agents; for this suite we select a set of representative cooperative environments. 

Following prior work~\cite{smac, marie}, we report the evaluation median and standard deviation of each experiment over four random seeds. Evaluation is based on the mean win rate or mean reward of all agents across 50 episodes. For SMAC, we compare against three world models (MARIE, MAMBA, MBVD) and three model-free baselines across 12 maps (9 \textit{easy}, 3 \textit{hard}). \textit{Easy} maps are trained for 50K steps, and \textit{hard} maps for 200K. While 100K is the standard for \textit{easy} maps, prior work has shown that many models already reach 100\% win rate under that budget \cite{marie, MAMBA}, so we use 50K to mitigate multi-way ties~\cite{marie, MAMBA}. For PettingZoo and MeltingPot, we compare against four model-free methods, using a 50K-step training budget. Hyperparameter settings for this study are provided in \ref{hyperparams}. 

We compare MATWM against several state-of-the-art baselines in MARL. First, we evaluate it against three world model architectures: MARIE, MAMBA, and MBVD, all of which represent the current frontier in world models for multi-agent systems. In addition to these model-based baselines, we also compare MATWM with a set of strong model-free algorithms, providing a broad and representative benchmark across different learning paradigms. QMIX \cite{Qmix} is a popular value-decomposition method that factors the joint action-value function into individual agent utilities under a monotonicity constraint. QPLEX \cite{QPLEX} extends QMIX by incorporating a duplex dueling architecture that enables more expressive value functions. MAPPO \cite{mappo} is a centralized variant of Proximal Policy Optimization \cite{ppo} adapted for multi-agent settings, using a shared critic across agents. MAA2C \cite{maa2c} (Multi-Agent Advantage Actor-Critic) is an early actor-critic method for MARL that uses decentralized policies with a shared critic to stabilize training. We do not include MAA2C in SMAC or other vector-based benchmarks, as we already compare against several strong model-free and model-based baselines. MAA2C is included in our image-based evaluations only (PettingZoo and MeltingPot) primarily to provide an additional relevant model-free baseline, since existing world model approaches are omitted from these studies. These model-free baselines are selected as they are widely-used strong performers across a variety of MARL benchmarks.

\subsection{SMAC Results}
\begin{figure}[H]
  \centering
  \includegraphics[width=\textwidth]{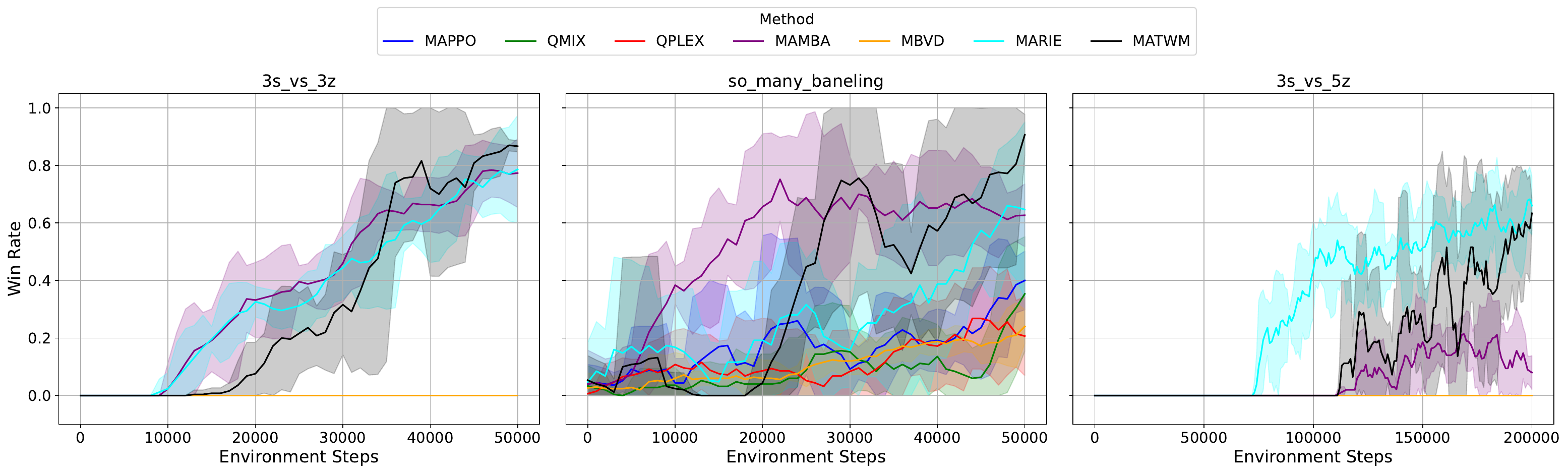}
  \caption{Curves of evaluation win rates for each method on three representative SMAC maps. Curves are smoothed for visual clarity of the graphs. The x-axis represents the number of steps taken in the environment, while the y-axis shows the corresponding evaluation win rate.}
  \label{fig:win_rates}
\end{figure}

\begin{table}[H]
\centering
\caption{Win rate as a \% comparison across various SMAC maps: Median (Std).}
\resizebox{\textwidth}{!}{
\begin{tabular}{l|l|cc|cc|cc|cc|cc|cc|cc}
\hline

\textbf{Map} & \textbf{Steps} 
& \multicolumn{2}{c|}{\textbf{MATWM}} 
& \multicolumn{2}{c|}{\textbf{MARIE}} 
& \multicolumn{2}{c|}{\textbf{MAMBA}} 
& \multicolumn{2}{c|}{\textbf{MBVD}} 
& \multicolumn{2}{c|}{\textbf{MAPPO}} 
& \multicolumn{2}{c|}{\textbf{QMIX}} 
& \multicolumn{2}{c}{\textbf{QPLEX}} \\
 & & \textbf{Med} & \textbf{Std} 
& \textbf{Med} & \textbf{Std} 
& \textbf{Med} & \textbf{Std} 
& \textbf{Med} & \textbf{Std} 
& \textbf{Med} & \textbf{Std} 
& \textbf{Med} & \textbf{Std} 
& \textbf{Med} & \textbf{Std} \\
\hline
2m\_vs\_1z      & 50K &  \textbf{98.0} & 3.2 & 95.0 & 4.4 & 91.0 & 6.2 & 41.0 & 20.7 & 51.0 & 10.3 & 62.0 & 8.9 & 70.0 & 13.0 \\
2s\_vs\_1sc     & 50K &  \textbf{96.0} & 5.7 & 90.0 & 9.1 & 80.0 & 7.3 & 0.0 & 1.2 & 18.0 & 7.6 & 8.0 & 4.4 & 13.0 & 6.8 \\
2s3z           & 50K &  \textbf{80.0} & 9.0 & 71.0 & 8.6 & 68.0 & 12.1 & 28.0 & 17.5 & 13.0 & 3.0 & 17.0 & 4.3 & 36.0 & 5.8 \\
3m             & 50K &  \textbf{83.0} & 10.4 & 78.0 & 14.1 & 68.0 & 7.7 & 60.0 & 9.2 & 54.0 & 6.3 & 61.0 & 10.2 & 58.0 & 4.9 \\
3s\_vs\_3z      & 50K &  \textbf{87.0} & 19.4 & 85.0 & 21.8 & 77.0 & 23.7 & 0.0 &0.0 & 0.0 & 0.0 & 0.0 & 0.0 & 0.0 & 0.0 \\
3s\_vs\_4z      & 50K &  \textbf{12.0} & 4.8 & 0.0 & 0.8 & 4.0 & 1.4 & 0.0 & 0.0 & 0.0 & 0.0 & 0.0 & 0.0 & 0.0 & 0.0 \\
8m             & 50K &  67.0 & 24.9 & \textbf{72.0} & 7.1 & 68.0 & 6.4 & 52.0 & 18.9 & 38.0 & 4.9 & 60.0 & 18.2 & 59.0 & 9.4 \\
MMM            & 50K &  \textbf{7.0} & 4.7 & 1.0 & 1.6 & 3.0 & 3.5 & 0.0 & 0.0 & 0.0 & 0.0 & 0.0 & 0.0 & 3.0 & 2.7 \\
so\_many\_baneling & 50K &  \textbf{86.0} & 22.9 & 73.0 & 12.4 & 66.0 & 14.2 & 27.0 & 12.3 & 31.0 & 7.6 & 40.0 & 6.1 & 52.0 & 8.8 \\
3s\_vs\_5z      & 200K &  64.0 & 26.5 & \textbf{66.0} & 28.0 & 6.0 & 10.1 & 0.0 & 0.0 & 0.0 & 0.0 & 0.0 & 0.0 & 0.0 & 0.0 \\
2c\_vs\_64zg    & 200K &  7.0 & 7.5 & \textbf{14.0} & 8.2 & 0.0 & 0.8 & 0.0 & 0.0 & 0.0 & 0.0 & 0.0 & 0.0 & 0.0 & 0.0 \\
5m\_vs\_6m      & 200K & \textbf{46.0} & 21.8 & 40.0 & 11.7 & 14.0 & 7.3 & 0.0 & 0.0 & 0.0 & 0.0 & 0.0 & 0.0 & 0.0 & 0.0 \\
\hline
\textbf{Mean} & — & \textbf{61.1} & — & 57.1 & — & 45.4 & — & 17.3 & — & 17.1 & — & 20.7 & — & 24.3 & — \\

\hline
\end{tabular}
}
\label{tab:smac}

\end{table}

We evaluate our framework on the StarCraft Multi-Agent Challenge, a widely adopted benchmark for multi-agent coordination under partial observability. Our model is compared against both model-free and model-based baselines across a diverse set of 12 scenarios, including nine \textit{easy} maps and three \textit{hard} maps. The results of these experiments are displayed in Table \ref{tab:smac}. We observe that, given a very small sample budget, MATWM outperforms all model-free baselines and even MBVD by a substantial margin, while also outperforming MAMBA and MARIE by lesser, but still decisive margins. On several \textit{easy} maps, such as 2m\_vs\_1z, 2s\_vs\_1sc, 2s3z, and so\_many\_baneling, our model exhibits near-optimal behavior, achieving win rates of 85\% or higher. In 3s\_vs\_3z, where tighter coordination is required, MATWM remains ahead, while model-free baselines fail to learn a winning strategy in the 50K sample budget. In 3s\_vs\_4z and MMM, the hardest of the \textit{easy} maps, it is difficult for even the world model baselines to consistently learn a winning strategy within the sample budget; MATWM, however, is able to do so more consistently than other world model baselines. Particularly in the \textit{hard} maps, such as 2c\_vs\_64zg and 3s\_vs\_5z, model-free methods exhibit no success, unable to learn a winning strategy, while MATWM continues to be able to do so consistently. MARIE shows slight gains over our method on certain \textit{hard} maps when given more training, but our model remains more sample efficient overall. Figure \ref{fig:win_rates} shows evaluation win rates measured every 1,000 environment steps across three representative SMAC maps. Compared to other world model baselines, MATWM often requires more time to learn a winning strategy. However, once it does, it rapidly refines and exploits that strategy, achieving significantly higher win rates in a shorter span. These results establish our framework as a leading model in low-data, multi-agent settings, especially where coordination and partial observability are critical.

\subsection{PettingZoo \& MeltingPot}
\begin{figure}[H]
  \centering
  \includegraphics[width=\textwidth]{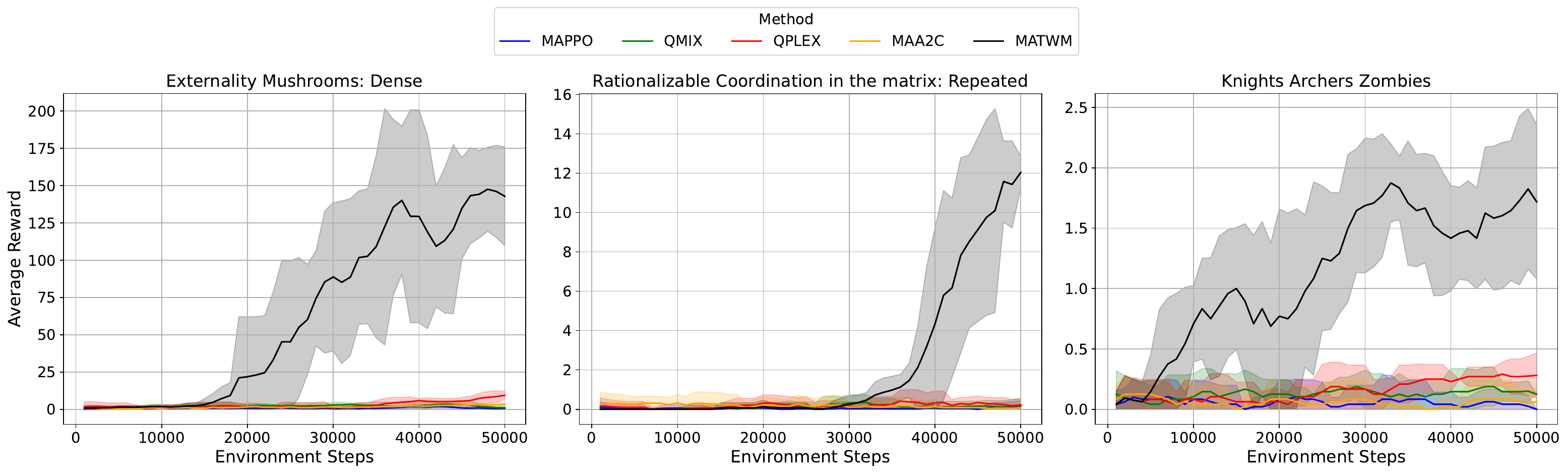}
  \caption{Curves of average reward obtained across all agents for each method on three representative PettingZoo and MeltingPot environments. Curves are smoothed for visual clarity of the graphs. The x-axis represents the number of steps taken in the environment, while the y-axis shows the corresponding average reward across all agents.}
  \label{fig:image_curves}
\end{figure}

\begin{table}[H]
\centering
\caption{Performance comparison across various PettingZoo Butterfly environments: Median (Std).}
\resizebox{\textwidth}{!}{
\begin{tabular}{l|l|cc|cc|cc|cc|cc}

\textbf{Environment} & \textbf{Steps} 
& \multicolumn{2}{c|}{\textbf{MATWM}} 
& \multicolumn{2}{c|}{\textbf{MAA2C}} 
& \multicolumn{2}{c|}{\textbf{MAPPO}} 
& \multicolumn{2}{c|}{\textbf{QMIX}} 
& \multicolumn{2}{c}{\textbf{QPLEX}} \\
 & & \textbf{Med} & \textbf{Std} 
& \textbf{Med} & \textbf{Std} 
& \textbf{Med} & \textbf{Std} 
& \textbf{Med} & \textbf{Std} 
& \textbf{Med} & \textbf{Std} \\
\hline
Cooperative Pong         & 50K &  \textbf{54.5} & 25.1 & 9.5 & 3.1 & 5.5 & 1.8 & 10.2 & 2.4 & 11.7 & 2.9 \\
Knights Archers Zombies     & 50K & \textbf{1.75} & 1.3  & 0.0 & 0.1 & 0.0 & 0.2 & 0.12 & 0.3 & 0.25 & 0.3 \\
Pistonball           & 50K &  \textbf{92.6} & 2.3  & 2.5 & 0.0 & -1.3 & 0.0 & 4.1 & 0.0 & 8.4 & 0.0 \\

\end{tabular}
}
\label{tab:pettingzoo}

\end{table}

\begin{table}[H]
\centering
\caption{Performance comparison across various MeltingPot Environments: Median (Std).}
\resizebox{\textwidth}{!}{
\begin{tabular}{l|l|cc|cc|cc|cc|cc}

\textbf{Environment} & \textbf{Steps} 
& \multicolumn{2}{c|}{\textbf{MATWM}} 
& \multicolumn{2}{c|}{\textbf{MAA2C}} 
& \multicolumn{2}{c|}{\textbf{MAPPO}} 
& \multicolumn{2}{c|}{\textbf{QMIX}} 
& \multicolumn{2}{c}{\textbf{QPLEX}} \\
 & & \textbf{Med} & \textbf{Std} 
& \textbf{Med} & \textbf{Std} 
& \textbf{Med} & \textbf{Std} 
& \textbf{Med} & \textbf{Std} 
& \textbf{Med} & \textbf{Std} \\
\hline

Chicken in the matrix: Arena         & 50K &  \textbf{21.5} & 2.4 & 1.3 & 0.8 & 2.1 & 1.1 & 3.1 & 1.4 & 2.7 & 0.9 \\
Coop Mining         & 50K &  \textbf{19.0} & 4.1 & 9.5 & 6.1 & 1.4 & 0.7 & 1.0 & 0.6 & 1.7 & 0.5 \\
Externality Mushrooms: Dense      & 50K &  \textbf{146.8} & 18.5  & 1.9 & 3.1 & 1.0 & 3.2 & 2.2 & 6.7 & 9.2 & 9.1 \\
Gift Refinements        & 50K &  \textbf{75.0} & 25.1 & 9.5 & 6.1 & 6.5 & 10.8 & 14.2 & 16.4 & 23.7 & 21.9 \\
Pure Coordination in the matrix: Repeated     & 50K & \textbf{6.8} & 0.7  & 0.0 & 0.4 & 0.0 & 0.0 & 0.0 & 0.0 & 0.0 & 0.1 \\
Rationalizable Coordination in the matrix: Repeated           & 50K &  \textbf{12.2} & 0.5  & 0.3 & 0.9 & 0.0 & 0.5 & 0.2 & 0.3 & 0.4 & 0.8 \\
Stag Hunt in the matrix: Arena           & 50K &  \textbf{7.2} & 2.5  & 1.0 & 0.7 & 0.7 & 0.4 & 0.0 & 0.7 & 1.2 & 1.0 \\
\end{tabular}
}
\label{meltingpot}
\end{table}

We evaluate the performance of MATWM on image-based multi-agent environments using the PettingZoo and MeltingPot benchmark suites. World model baselines are not included for PettingZoo and MeltingPot due to the absence of published adaptations or results for use in image-based environments. To the best of our knowledge, MATWM is the first world model framework evaluated in this domain. The results in Tables \ref{tab:pettingzoo} and \ref{meltingpot} highlight the strong generalization and sample efficiency of our model in visually complex settings. In the PettingZoo scenarios, MATWM consistently outperforms all four model-free baselines across all evaluated tasks. For instance, despite Pistonball's staggering 20 agents, MATWM is still able to achieve near-optimal performance with a very small sample budget, outperforming model-free baselines by a significant margin. We observe a similar trend in Cooperative Pong and Knights Archers Zombies, a substantial margin over baselines, whereas the model-free baselines perform roughly similar to a random policy. This showcases our model’s capacity to learn effective cooperative behaviors from visual observations under tight sample constraints. MeltingPot results further reinforce these trends. Our model consistently achieves high rewards across all scenarios, regardless of the number of agents, which range between two and eight, compared to scores from all baselines. Even in more subtle or sparse-reward environments like Pure Coordination in the matrix: Repeated, MATWM is still able to quickly learn strategies that achieve high rewards. Similarly to the PettingZoo experiments, model-free baselines behave at or only slightly better than a random policy on nearly all evaluated environments, highlighting their inability to learn meaningful behavior under limited training budgets. Previous studies conducted on PettingZoo Butterfly environments suggest that it would take nearly one million or more environment steps for any of the model-free baselines to achieve comparable results to what MATWM did in just 50k \cite{extbenchmark, marlbenchmark}. Figure \ref{fig:image_curves} shows performance curves for MATWM and baseline methods evaluated at after every 1000 environment steps across three representative environments: one from PettingZoo and two from MeltingPot. These results demonstrate the benefits of incorporating imagined trajectories for sample efficiency as baselines often fail to make progress. One recurring pattern we observed during training, also visible in the figure, is that MATWM's performance occasionally dips before stabilizing again. Although this may be environment dependent, we hypothesize that such fluctuations may actually stem from the world model temporarily falling behind the rapidly updating agent policies. This mismatch can momentarily destabilize training and reflects a known limitation of the decentralized world model approach, even when using prioritized replay buffers. Nonetheless, MATWM’s ability to handle image-based inputs, unlike prior multi-agent world models, broadens its applicability and represents a significant advancement toward general-purpose, sample efficient multi-agent learning.

\subsection{Ablation Study}

To evaluate the contribution of key architectural components in MATWM, we perform an ablation study on three core mechanisms: the teammate predictor, prioritized experience replay (PER), and action scaling. For each component, we disable the respective mechanism and measure the performance drop across a representative subset of SMAC maps and image-based environments. The results are summarized in Table~\ref{tab:component-ablation}.

\textbf{Teammate Predictor.}  
Disabling the teammate predictor consistently leads to substantial performance degradation, particularly in environments involving many agents and requiring coordinated behavior (e.g., 8m, so\_many\_baneling), where performance often collapses to zero. In smaller-scale or loosely coupled scenarios (e.g., 3m, Pure Coordination in the matrix: Repeated), the model without the predictor still performs reasonably, though it remains outperformed by the full model. In the case of Cooperative Pong, we observe that the teammate predictor provides little to no benefit, and in some runs, even hinders performance. We hypothesize that this is because coordination is not critical in this environment; the primary focus is on tracking the ball, and the actions of the other agent during gameplay may introduce noise or distraction rather than useful information. However, this study confirms that explicitly modeling teammates' actions significantly enhances cooperation in tightly coupled environments.

\textbf{Prioritized Experience Replay.}  
Replacing PER with uniform sampling results in a noticeable decline in performance, particularly in environments where agents take longer to learn a winning strategy, such as 8m. In such settings, agent behaviors converge more slowly, meaning the replay buffer contains a greater proportion of outdated or suboptimal experiences. PER mitigates this by sampling transitions that are more recent, which allow the world model to track and update with agents' changing behavior. This ensures that updates focus on more relevant experiences, improving imagination and thus sample efficiency. In contrast, in simpler environments like 3m, where agents quickly discover successful strategies and thus stop rapidly changing behavior, the replay buffer is filled with relevant experiences early on, reducing the relative benefit of prioritization.

\textbf{Action Scaling.}  
Removing the action scaling mechanism also impacts performance, though to a lesser extent than the other ablations. The degradation is most evident in image-based environments, such as Pistonball and Externality Mushrooms: Dense, which can involve multiple agents being present on the screen simultaneously, making it crucial for the world model to identify which agent is conducting the action. We hypothesize that action scaling is particularly beneficial in fully observable settings, where the observation includes the entire environment state and agent differentiation is not implicitly encoded. In contrast, in partially observable environments like SMAC, the focal agent’s identity is often loosely embedded within its localized observation as the focal agent's information is represented by the first values in the vector, allowing the model to maintain some degree of agent distinction even without scaling. In environments like Cooperative Pong, however, where the observation encompasses the whole board, scaling the action distribution helps the model maintain distinct agent behaviors, improving the learning stability of the world model.

Together, these results highlight that MATWM's performance emerges not solely from its predictive world model, but from the synergy between multiple design decisions—including teammate prediction, prioritized sampling, and consistent action scaling.

\begin{table}[h]
\centering
\caption{Ablation Study: Impact of Component Removal on Performance}
\label{tab:component-ablation}
\resizebox{\linewidth}{!}{%
\begin{tabular}{lcccc}
\toprule
\textbf{Environment} & \textbf{Full MATWM} & \textbf{-- Teammate Predictor} & \textbf{-- PER} & \textbf{-- Action Scaling} \\
\midrule
3m                                 & 83.0 & 65.0 & 74.0 & \textbf{84.0} \\
8m                                 & \textbf{65.0} & 0.0  & 52.0 & 63.0 \\
so\_many\_baneling                 & \textbf{74.0} & 0.0  & 59.0 & 70.0 \\
Cooperative Pong                         & 54.5 & \textbf{56.3} & 52.1 & 38.4 \\
Pistonball                         & \textbf{92.6} & 90.3 & 85.1 & 88.4 \\
Externality Mushrooms: Dense       & \textbf{146.8} & 130.0 & 133.5 & 135.7 \\
Pure Coordination in the matrix: Repeated & \textbf{6.8} & 5.9 & 6.0 & 6.5 \\
\bottomrule
\end{tabular}%
}
\end{table}

\section{Conclusion}
\label{sec:conclusion}

We introduce MATWM, a transformer-based multi-agent world model that builds upon STORM, a leading world model architecture for single-agent systems, which achieves strong performance due to its aggregation of many state-of-the-art techniques and architectural advances. MATWM uses imagination-based training, a decentralized world model architecture, and a novel teammate predictor to facilitate coordination in partially observable environments. We also incorporated several elements from state-of-the-art single-agent world models that have not yet been applied to multi-agent world models, such as symlog two-hot rewards, KL balance and free bits, and percentile return normalization. 
MATWM achieves state-of-the-art performance across a diverse suite of benchmarks, including SMAC, PettingZoo, and MeltingPot, demonstrating strong sample efficiency in complex multi-agent settings by obtaining near-optimal performance in many of these benchmarks in as few as 50K environment interactions, the smallest budget attempted in existing MARL literature. Our experiments highlight the benefits of architectural choices such as prioritized experience replay, action scaling for agent disambiguation, and the explicit modeling of teammate behavior. Notably, we show that MATWM excels not only in traditional vector-based benchmarks but also in complex image-based environments, whereas previous multi-agent world models were limited to vector-based environments.

We hope MATWM provides a solid foundation for future research into more advanced architectures capable of addressing increasingly complex challenges in multi-agent reinforcement learning. In future work, we aim to enhance our framework by investigating additional mechanisms for facilitating multi-agent cooperation, extending support to competitive and mixed cooperative-competitive MARL settings, and addressing limitations in decentralized world models that are related to non-stationarity.

\section*{Declaration of Generative AI and AI-Assisted Technologies in the Writing Process}

During the preparation of this work, the author(s) used ChatGPT (OpenAI) in order to support language editing. After using this tool, the author(s) reviewed and edited the content as needed and take(s) full responsibility for the content of the publication.

\newpage
\bibliographystyle{elsarticle-num}
\bibliography{bib}

\newpage
\appendix
\section{Imagination Algorithm}
\label{imaginationalgo}
\begin{algorithm}[H]
\caption{Imagined Rollouts with Teammate Prediction}
\label{alg:multiagent_imagination}
\DontPrintSemicolon
\KwIn{Agents $\mathcal{N}$, world model $WM$, context observations/actions, context length $L$, batch size $B$, imagination rollout length $H$}
\KwOut{Imagined latent states, actions, rewards, terminations, teammate actions, action masks}

Initialize per-agent rollout buffers for latent states, actions, rewards, terminations, teammate actions, action masks (optional)\;

\tcp{Use Context as starting point for imagination rollout}
Encode context observations into latent state sequence $z_{0:L}$\; 
\For{$t = 0$ \KwTo $L-1$}{
    Predict next latent state, hidden state, reward, and termination using:\;
    \quad $(\hat{z}_{t+1}, h_{t+1}, \hat{r}_t, \hat{c}_t) \gets WM(z_t, a_t)$\;
}
Store $z_L$ and hidden state $h_L$ in respective buffers\;
\tcp{Begin imagination rollout}
\For{$t = 0$ \KwTo $H-1$}{ 
    \ForEach{agent $n_i \in \mathcal{N}$}{
        Get latent $z_t^{(i)}$ and hidden $h_t^{(i)}$ slice for agent batch\;
        Compute predicted teammate action probabilities: $\hat{p}_{i, t}^{(a)} \gets \text{TeammatePredictor}(z_t^{(i)})$\;

        \uIf{action masking enabled}{
            Predict action mask $\hat{m}_t \gets \text{MaskPredictor}(z_t^{(i)})$\;
            Sample action $a_{i,t} \sim \pi(z_t^{(i)}, h_t^{(i)}, \hat{p}_{i, t}^{(a)}, \hat{m}_t)$\;
            Store $\hat{m}_t$ in respective buffer
        }\Else{
            Sample action $a_{i,t} \sim \pi(z_t^{(i)}, h_t^{(i)}, \hat{p}_{i, t}^{(a)})$\;
        }

        Store $a_{i,t}$ and $\hat{p}_{i, t}^{(a)}$ in respective buffers\;
    }
    Predict next $(\hat{z}_{t+1}, h_{t+1}, \hat{r}_t, \hat{c}_t) \gets WM(z_t, a_t)$\;
    Store results in respective buffers\;
}

Return buffers for latent states, actions, rewards, terminations, teammate actions, and (optional) masks\;

\end{algorithm}
\section{Training Algorithm}
\label{matwmalgo}
\begin{algorithm}[H]
\caption{Joint Training of World Model and Agents in MATWM}
\label{alg:matwm}
\DontPrintSemicolon
\KwIn{Environment $\mathcal{E}$, world model $WM$, actor-critic agents $\{\pi_i\}_{i=1}^{N}$, replay buffers $\{RB_i\}_{i=1}^{N}$}
\KwParams{Total steps $T$, train intervals $t_{WM}, t_{\pi}$, batch sizes, context lengths}

Initialize environment $\mathcal{E}$ and per-agent buffers, context queues\;
\For{$t = 1$ \KwTo $T$}{
    \ForEach{agent $i$}{
        \eIf{$RB_i$ is ready}{
            Encode recent observations $\mathcal{O}_{i}$ into latent $z_i$ using $WM$\;
            Predict teammate behavior using $WM$'s predictor\;
            Sample action $a_i \sim \pi_i(z_i, \hat{a}_{-i})$\;
        }{
            Sample $a_i$ randomly or with action masking\;
        }
    }
    Step environment $\mathcal{E}$ with $\{a_i\}$, collect next obs, rewards, dones\;
    Store transitions in each $RB_i$ with teammate action info\;

    \If{episode terminates}{
        Reset environment and context buffers\;
    }

    \If{$t \bmod t_{WM} = 0$ and all $RB_i$ ready}{
        Sample batches from all $RB_i$\;
        Train $WM$ on combined batch (obs, actions, rewards, terminations, teammate actions)\;
    }

    \If{$t \bmod t_{\pi} = 0$ and all $RB_i$ ready}{
        Use $WM$ to imagine trajectories for each agent\;
        Train each policy $\pi_i$ using imagined data (rewards, terminations, predicted teammates)\;
    }

}
\end{algorithm}

\section{Reproducibility}
\label{hyperparams}
Code for MATWM is available on GitHub \footnote{github.com/azaddeihim/matwm}. For all SMAC experiments, we use Starcraft II version SC2.4.1.2.60604. Please note that using different versions may yield slightly different results. Hyperparameter settings for MATWM are shown in Table \ref{tab:matwm-hyperparams}. The settings for all baselines are taken from \cite{marie} for the SMAC experiments and \cite{extbenchmark} for the PettingZoo and MeltingPot experiments. 

\begin{table}[h]
\centering
\caption{MATWM Hyperparameter Settings}
\label{tab:matwm-hyperparams}
\begin{tabular}{l c}
\toprule
\textbf{Hyperparameter} & \textbf{Value} \\
\midrule
Max sequence length & 64 \\
Hidden dimension & 512 \\
Number of layers & 2 \\
Number of attention heads & 8 \\
Latent dimension size & 32 \\
Number of categories per latent & 32 \\
World model train batch length & 64 \\
World model train batch size & 16 \\
World Model Learning rate & $3 \times 10^{-5}$ \\
Actor+Critic Learning rate & $3 \times 10^{-4}$ \\
Optimizer & Adam \\
Gradient clipping agent & 100.0 \\
Gradient clipping world model & 1000.0 \\
Replay buffer size & 50,000  \\
Replay sampling priority decay & 0.9998 \\
KL loss weight ($\beta_1$) & 0.5 \\
Representation loss weight ($\beta_2$) & 0.1 \\
Imagination horizon & 16 \tablefootnote{\label{fn}These settings are for environments with three or less agents. For environments with four to six agents and for 2c\_vs\_64z, we use a batch size of 768 with an imagination horizon of 12. For environments with seven or more agents, we use a batch size of 1024 with an imagination horizon of 8.} \\
Agent train batch size & 512$^{\text{\ref{fn}}}$ \\
Imagination context length & 8 \\
Train world model every $n$ steps & 1 \\
Train agent every $n$ steps & 1 \\
MLP Encoder Hidden dim & 512 \\
MLP Encoder Hidden layers & 3 \\
CNN Encoder Kernel Size &  4 \\
CNN Encoder Stride & 2 \\
CNN Encoder Layers & 4 \\
\bottomrule
\end{tabular}
\end{table}
\end{document}